\documentclass[conference]{IEEEtran}
\IEEEoverridecommandlockouts
\usepackage{subcaption}
\usepackage{tikz}
\usepackage{cite}
\usepackage{amsmath,amssymb,amsfonts}
\usepackage{algorithmic}
\usepackage{graphicx}
\usepackage{textcomp}
\usepackage{xcolor}
\usepackage{balance}
\def\BibTeX{{\rm B\kern-.05em{\sc i\kern-.025em b}\kern-.08em
    T\kern-.1667em\lower.7ex\hbox{E}\kern-.125emX}}

\begin{document}

\title{A Corridor-Scale CARLA–VISSIM Co-Simulation Framework for Multi-Intersection Urban Traffic}


\author{
\IEEEauthorblockN{Sima Ashayer, Austin Haris, Mina Sartipi}
\IEEEauthorblockA{The University of Tennessee at Chattanooga, Chattanooga, TN, USA\\
npk875@mocs.utc.edu, austin-p-harris@utc.edu, mina-sartipi@utc.edu}
}

\maketitle
\vspace{-2mm}

\begin{abstract}
This paper presents an implemented CARLA–VISSIM co-simulation framework for an urban corridor comprising approximately fifteen connected intersections centered on Martin Luther King Jr. Boulevard in Chattanooga, Tennessee. The system integrates CARLA~0.10.0 Unreal Engine 5 with PTV VISSIM 2026 through a bidirectional, step-synchronized interface that couples VISSIM’s microscopic vehicle, pedestrian, and signal-controller logic with CARLA’s high-fidelity 3D rendering. A LiDAR-derived elevation model and RoadRunner-based High Definition (HD) map provide terrain-accurate road geometry deployed consistently across both simulators. The framework incorporates explicit actor ownership, mirrored lifecycle management, coordinate reconciliation, and a latest-state-per-actor update policy, enabling stable interaction between VISSIM-controlled traffic and a CARLA-controlled ego vehicle. A corridor-scale case study demonstrates consistent traffic-signal mirroring, synchronized vehicle–pedestrian interactions, and stable mixed-authority operation under peak loads of approximately 100 vehicles and 100 pedestrians. The deployment captures the interaction of the five signalized intersections along MLK Street and their connecting upstream and downstream intersections, revealing synchronization challenges unique to multi-intersection corridors. Results indicate that this MLK-centered corridor provides an effective testbed for verifying cross-simulator consistency and that the proposed architecture supports reliable, perception-ready co-simulation for corridor-level traffic studies.
\end{abstract}


\section{Introduction}
Intelligent transportation systems require accurate modeling of vehicle–pedestrian interactions, multimodal traffic flow, and signalized corridor operations. Because real-world testing of connected and autonomous vehicles (CAVs) is costly, risky, and unable to capture the full range of vehicle-to-everything (V2X) behaviors, simulation has become essential for evaluating dense urban interactions. Smart-corridor research depends on calibrated microscopic traffic dynamics and high-fidelity 3D environmental representations that can jointly represent detailed multimodal interactions in realistic settings \cite{ashayer2025co}.

Microscopic traffic simulators such as PTV VISSIM and SUMO are widely used in transportation research to model traffic flow, pedestrian behavior, and signal control \cite{PTVVissim, lopez2018microscopic}. While these tools capture traffic operations and control logic, they lack realistic 3D environments and sensor-level perception needed to evaluate perception- and behavior-level performance of CAVs. In contrast, high-fidelity driving simulators such as CARLA provide photorealistic environments and physics-based sensor simulation, enabling perception-oriented studies and interactive driving experiments \cite{dosovitskiy2017carla}. However, CARLA does not natively support microscopic traffic modeling or signal control comparable to dedicated traffic simulators. As a result, integrating microscopic traffic simulation with high-fidelity driving environments through co-simulation has emerged as a practical approach for realistic urban traffic evaluation. While CARLA–SUMO co-simulation is widely studied\cite{azfar2025traffic}, this work adopts VISSIM for its practitioner-grade signal-controller modeling and detailed multimodal behavior. These capabilities are widely used in transportation engineering practice and are essential for corridor-scale evaluation.

This paper presents a corridor-scale \footnote{In this paper, the term corridor-scale refers to the MLK Smart corridor in downtown Chattanooga, modeled with its real road geometry and elevation} co-simulation framework that integrates CARLA (UE5) and PTV VISSIM-2026 under a bidirectional, step-synchronized architecture. The framework incorporates LiDAR-derived elevation maps created from public United States Geological Survey (USGS) data, supports bidirectional vehicle-state exchange, and supplies all pedestrian trajectories and signal logic from VISSIM to CARLA. A fifteen-intersection segment of Chattanooga’s urban network, comprising five signalized and ten stop-controlled intersections, is used as a case study to demonstrate the framework’s applicability for smart-corridor evaluation.

\section{Related works}
Several studies have explored co-simulation frameworks that integrate CARLA with SUMO. These CARLA–SUMO systems synchronize a microscopic traffic engine with a high-fidelity 3D driving simulator for applications such as traffic signal control, terrain-aware traffic simulation, and multi-simulator experimentation \cite{azfar2025traffic,raskoti2025elevation,ahmad2025opencams}. These works illustrate ongoing research on CARLA–SUMO co-simulation for diverse application objectives, but do not address multi-intersection corridor deployments or provide detailed vehicle, pedestrian, and signal-controller behavior at corridor scale, which motivates the use of VISSIM for corridor-level co-simulation.

Beyond CARLA–SUMO systems, a smaller set of studies has explored co-simulation between CARLA and VISSIM. One work implements a multimodal CARLA–VISSIM system integrating drivers, bicyclists, and pedestrians with traffic signal synchronization modules using real-time data exchange and custom scripts to study dynamic human responses in shared environments \cite{erdagi2025development}. Another study uses CARLA and VISSIM with high-definition maps to assess autonomous vehicle safety, but its application is limited to a few intersections and does not examine corridor-scale synchronization or bidirectional actor management \cite{li2025utilizing}. These CARLA–VISSIM approaches do not address the architectural and scalability challenges of multi-intersection corridor co-simulation. The present work instead emphasizes corridor-scale synchronization with explicit actor ownership, synchronous lockstep execution, and coordinated multi-intersection traffic operation.

Prior works show the value of combining microscopic traffic simulation with high-fidelity driving simulators, but several gaps remain for smart-corridor research. Existing systems operate in UE~4, while UE~5 introduces improved rendering via Nanite virtualized geometry and Lumen global illumination for more realistic perception-driven co-simulation \cite{epicUE5Nanite,epicUE5Lumen,turkcan2024boundlessgeneratingphotorealisticsynthetic}. To date, no published CARLA VISSIM framework provides a reusable, implementation-level architecture for corridor-scale co-simulation, where vehicles, pedestrians, and signal logic are synchronized across many intersections. This work addresses these gaps through a corridor-scale CARLA-VISSIM co-simulation implemented in CARLA~0.10.0 for UE5.

The contributions of this work are as follows:
(1) a reusable corridor-scale co-simulation framework integrating CARLA~0.10.0~(UE5) and PTV VISSIM 2026 under a bidirectional, step-synchronized interaction model with pedestrian movements supplied from VISSIM;
(2) an explicit actor-ownership and lifecycle management approach supporting vehicles, pedestrians, and traffic signals across simulators;
(3) a synchronization architecture using latest-state-per-actor updates, coordinate reconciliation, and batched control for stable operation under high traffic density;
(4) a LiDAR-derived HD map deployed consistently in both simulators for realistic multi-intersection evaluation. A fifteen-intersection case study shows realism, stability, and mixed-authority capability.

\section{System Overview and Architecture}
This paper presents a reusable corridor-scale co-simulation pipeline that integrates CARLA UE5 and VISSIM for smart-corridor experimentation in realistic urban environments. The framework enables co-simulation of vehicles, pedestrians, and traffic control systems by combining photorealistic sensor-level rendering with microscopic traffic modeling using LiDAR-derived elevation data and real road network topology to ensure behavioral realism and geometric fidelity.

\begin{figure}[!t]
\centering
\includegraphics[width=3.2in]{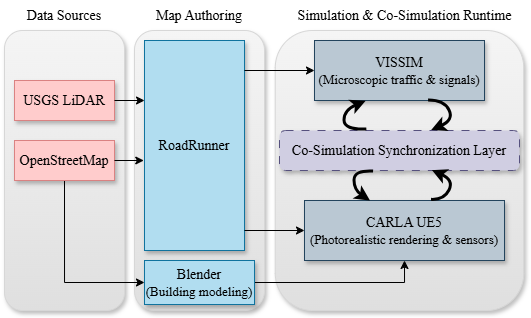}
\caption{\textbf{Overview of the CARLA–VISSIM co-simulation pipeline}. USGS LiDAR and OSM data generate a terrain-aware HD map in RoadRunner, while OSM building footprints support Blender-based building modeling for CARLA UE5. The finalized map is exported to both CARLA UE5 and VISSIM. VISSIM governs traffic flow, pedestrians, and signals, and CARLA provides high-fidelity rendering. A step-synchronized co-simulation layer enables bidirectional exchange of actor and signal states.}
\label{fig_architecture}
\end{figure}

Figure~\ref{fig_architecture} provides an overview of the system architecture. Publicly available USGS LiDAR data and OpenStreetMap (OSM) information are processed to generate a terrain-aware road network in RoadRunner. Road geometry is authored and refined in RoadRunner, while building assets are modeled in Blender using OSM data and imported into the UE, which serves as CARLA’s visual and sensor simulation backend. The resulting map preserves real-world terrain variation and is deployed consistently in both CARLA and VISSIM to ensure geometric alignment. Within this framework, VISSIM manages vehicle, pedestrian, and signal control logic, while CARLA (UE5) provides high-fidelity rendering and sensor simulation. This division allows each simulator to operate in its domain of strength while remaining tightly synchronized. 

A step-synchronized state-exchange mechanism maintains consistency between simulators. Vehicle states (position, orientation, and velocity) are exchanged bidirectionally at each timestep, while pedestrian states are supplied from VISSIM to CARLA to match the microscopic traffic model. Traffic-signal states are also transmitted from VISSIM to CARLA to keep visual and logical phases aligned. This mechanism ensures that traffic participants and control devices evolve coherently across both simulation environments during runtime. 

The framework supports corridor-scale deployments by enabling coordinated simulation across multiple connected intersections with realistic terrain changes, pedestrian facilities, and traffic control elements. By decoupling traffic logic and control from sensor and environment rendering, the proposed pipeline enables flexible experimentation for smart-corridor research, including vehicle–pedestrian interactions, traffic signal evaluation, and perception-driven studies. The modular structure of the framework allows the pipeline to be reused and extended to additional corridors or cities with minimal modification, supporting reproducibility and future expansion. 

\section{Data Sources and Preprocessing}
\noindent\textbf{LiDAR and Geospatial Data.}
Publicly available airborne LiDAR data from the U.S. Geological Survey (USGS) were used to model terrain elevation for the study corridor. A spatial subset corresponding to the corridor was extracted from a LiDAR tile of the USGS LiDAR Point Cloud (LPC) dataset for Hamilton County, Tennessee (2011 acquisition) \cite{usgs_national_map_lidar}. The dataset contains georeferenced three-dimensional point measurements. The LiDAR data were processed in MATLAB. Ground points were identified using a surface-based filtering approach suitable for terrain extraction \cite{mathworks_roadrunner_lidar}, and the resulting point cloud was used to derive elevation for map generation. Data were georeferenced using embedded coordinate reference system (CRS) information; when CRS metadata were incomplete, the spatial reference was inferred and validated to ensure consistency across datasets.

\noindent\textbf{Road Network Data.}
Road network topology was obtained from OSM \cite{OpenStreetMap}. Rather than relying on pre-downloaded files, OSM data were automatically retrieved based on the spatial extent of the LiDAR point cloud. The extracted data include road centerlines and intersection connectivity for the selected corridor. In addition to road topology, OSM data were used to obtain building footprints, and basic geometry outlines were later modeled in Blender and imported into the CARLA UE5 for visual realism, as part of the map construction pipeline.

\noindent\textbf{Preprocessing Overview.}
Preprocessing was limited to spatial filtering, coordinate alignment, and semantic processing of LiDAR points required to support subsequent HD map generation. No assumptions about lane geometry, traffic behavior, or simulation logic were introduced at this stage. Detailed map construction steps are described in the next section.

\section{HD Map Generation in RoadRunner}

\subsection{Road Network Reconstruction}
HD map generation was performed using MATLAB-based processing and RoadRunner’s HD map construction tools. OSM road data were automatically retrieved from the LiDAR spatial extent and imported into MATLAB, where road centerlines and intersection connectivity defined the initial network structure. The MATLAB driving scenario was then converted into a RoadRunner-compatible HD map format using built-in conversion utilities. This conversion preserves road centerlines and intersection topology, producing a centerline-based road geometry that serves as the scaffold for refinement in RoadRunner. LiDAR data were not used to introduce semantic objects or roadside assets at this stage. Instead, the point cloud was utilized to define the geographic extent of the environment and to support elevation integration, which is described in the following section. The resulting map serves as a unified representation of the road layout and forms a consistent foundation for deployment across simulation environments.

\subsection{Elevation and Terrain Modeling}
After road network reconstruction, LiDAR-derived elevation data were integrated to capture real-world roadway height variation. The georeferenced USGS point cloud was applied to the MATLAB road-network model using built-in elevation mapping utilities. Elevation integration assigns height values to the reconstructed road geometry based on the underlying point cloud, enabling the representation of longitudinal and lateral grade variations along the corridor. The resulting elevation-aware road network provides a more realistic geometric foundation for subsequent refinement and simulation deployment.

\subsection{Manual Map Refinement}
\begin{figure}[!t]
\centering
\includegraphics[width=2.3in]{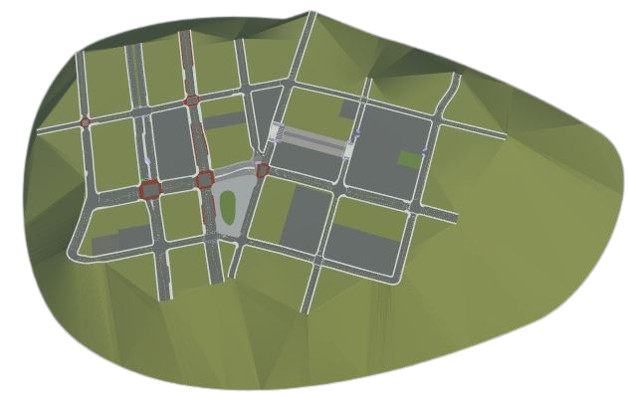}
\caption{\textbf{RoadRunner HD map of the selected urban corridor after elevation integration and refinement.} Lane geometry, intersection layouts, and terrain elevation are shown.}
\label{fig_RoadRunner}
\end{figure}

After automated road network reconstruction and elevation integration, the HD map was imported into RoadRunner for manual refinement to correct geometric inaccuracies not resolved through OSM conversion or elevation mapping. Refinements focused on lane alignment, intersection geometry, and network continuity, particularly where automated generation produced misaligned lanes, unrealistic turning paths, or discontinuities. These adjustments ensured smooth vehicle trajectories and consistent connectivity across the road network. Intersections and transition regions received particular attention because minor geometric inconsistencies can affect simulation stability. The finalized RoadRunner map was then exported to CARLA and VISSIM without further modification. A visualization of the refined network is shown in Figure~\ref{fig_RoadRunner}, which serves as the geometric reference for both simulators.

\section{Deployment in Simulation Environments}
\subsection{Development in CARLA (Unreal Engine 5)}
The finalized HD map generated in RoadRunner was deployed in CARLA as the base simulation environment. The map, including lane-level geometry, intersection layouts, and LiDAR-derived elevation, was imported to define the corridor's roadway infrastructure. To provide urban context beyond road geometry, building geometry was generated in Blender from OSM building footprints, and the resulting building shapes and positions were aligned with the RoadRunner road network. This approach enables consistent placement of urban structures while maintaining a lightweight representation suitable for large-scale simulation. The CARLA environment supports the visualization of dynamic entities such as vehicles, pedestrians, and traffic lights. The imported environment was configured to generate sensor data aligned with the deployed road geometry and urban context, enabling perception-oriented experimentation in a realistic urban setting.

\subsection{Visual and Environmental Enhancement in UE5}
\begin{figure*}[t]
\centering
    \begin{subfigure}[t]{0.4\textwidth}
        \centering
        \setlength{\fboxsep}{0pt}
        \setlength{\fboxrule}{0.1pt}
        \fbox{\begin{minipage}[c][1.5cm][c]{7cm}
            \includegraphics[width=7cm,height=1.5cm]{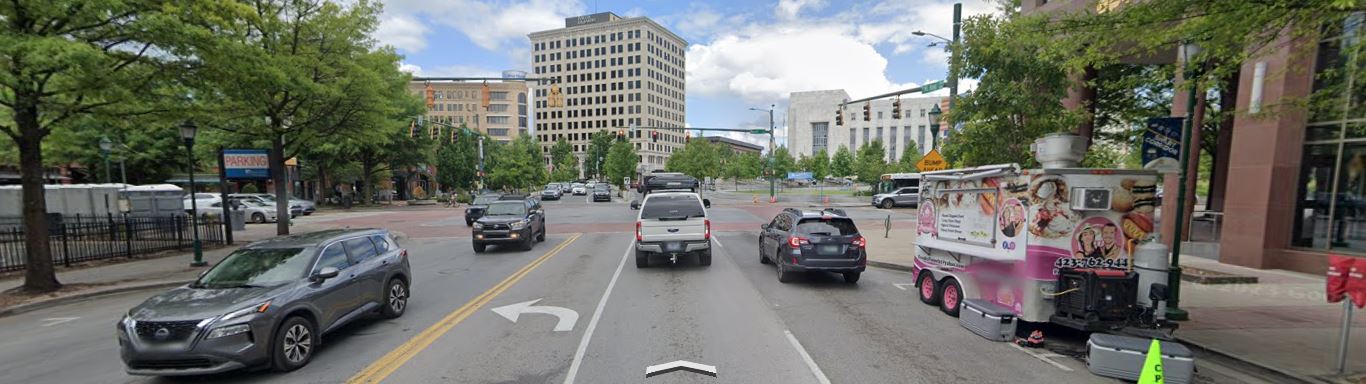}
        \end{minipage}}
    \end{subfigure}
    \begin{subfigure}[t]{0.4\textwidth}
        \centering
        \setlength{\fboxsep}{0pt}
        \setlength{\fboxrule}{0.1pt}
        \fbox{\begin{minipage}[c][1.5cm][c]{7cm}
            \includegraphics[width=7cm,height=1.5cm]{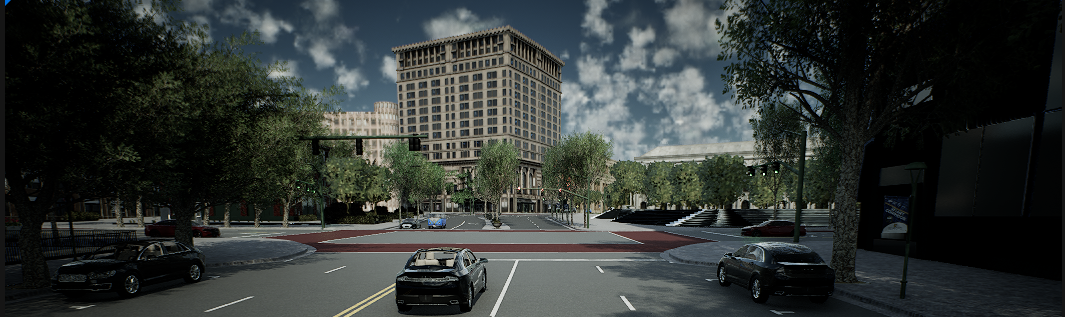}
        \end{minipage}}
    \end{subfigure}


     \begin{subfigure}[t]{0.4\textwidth}
        \centering
        \setlength{\fboxsep}{0pt}
        \setlength{\fboxrule}{0.1pt}
        \fbox{\begin{minipage}[c][1.5cm][c]{7cm}
            \includegraphics[width=7cm,height=1.5cm]{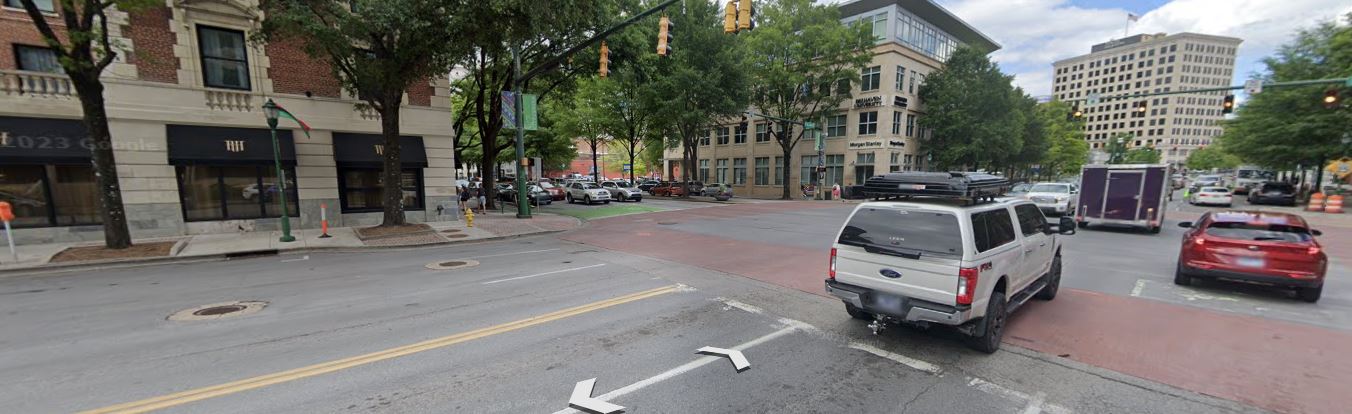}
        \end{minipage}}
    \end{subfigure}
    \begin{subfigure}[t]{0.4\textwidth}
        \centering
        \setlength{\fboxsep}{0pt}
        \setlength{\fboxrule}{0.1pt}
        \fbox{\begin{minipage}[c][1.5cm][c]{7cm}
            \includegraphics[width=7cm,height=1.5cm]{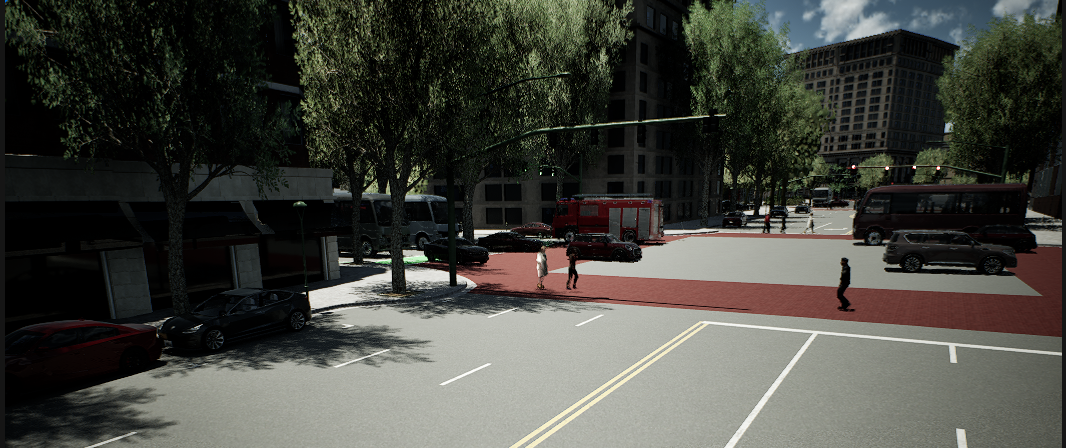}
        \end{minipage}}
    \end{subfigure}


    \begin{subfigure}[t]{0.4\textwidth}
        \centering
        \setlength{\fboxsep}{0pt}
        \setlength{\fboxrule}{0.1pt}
        \fbox{\begin{minipage}[c][1.5cm][c]{7cm}
            \includegraphics[width=7cm,height=1.5cm]{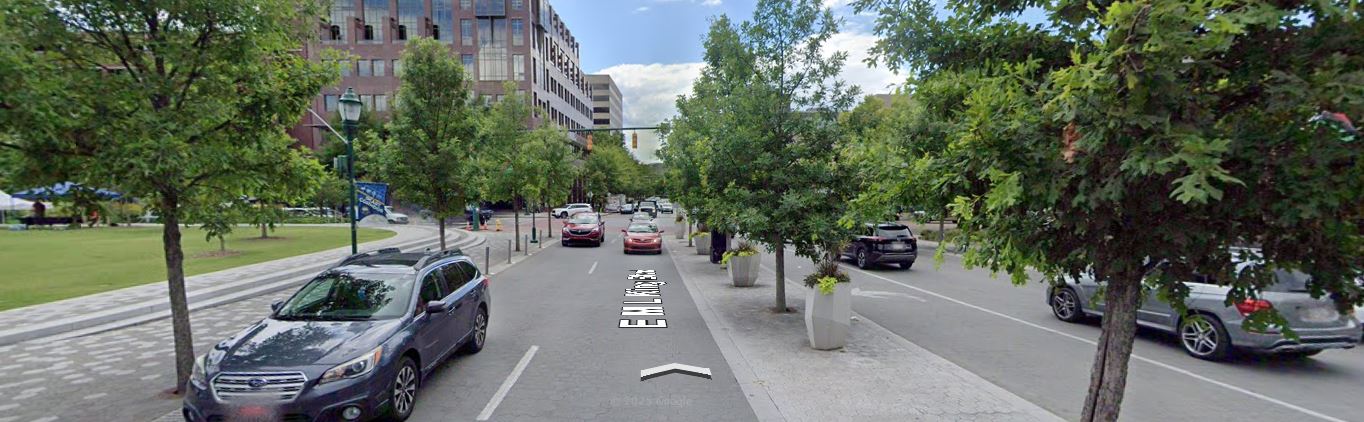}
        \end{minipage}}
    \end{subfigure}
    \begin{subfigure}[t]{0.4\textwidth}
        \centering
        \setlength{\fboxsep}{0pt}
        \setlength{\fboxrule}{0.1pt}
        \fbox{\begin{minipage}[c][1.5cm][c]{7cm}
            \includegraphics[width=7cm,height=1.5cm]{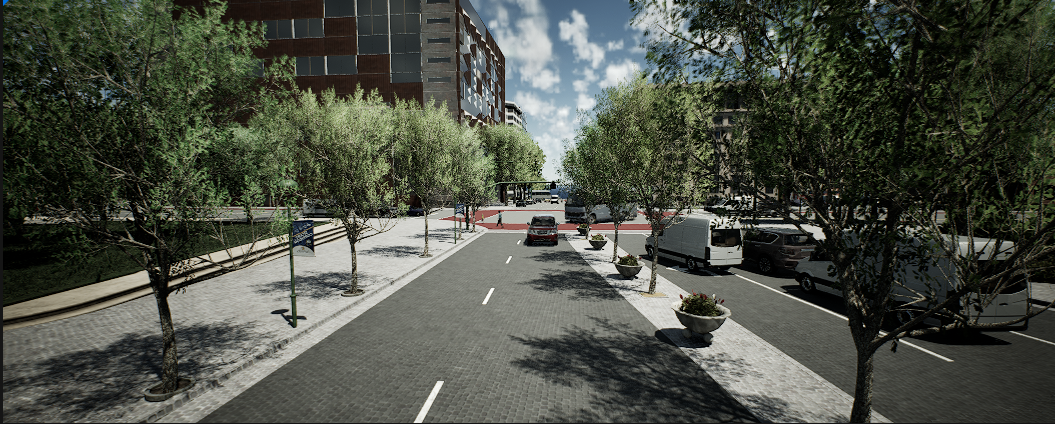}
        \end{minipage}}
    \end{subfigure}

\caption{\textbf{Side-by-side views of three representative locations along the study corridor in Google Earth (left) and the corresponding simulated scenes in CARLA UE5 (right).}}
\label{fig_visual_comparison}
\end{figure*}

Following deployment of the road network and building geometry, additional environmental adjustments were made in UE5 to enhance visual coherence and spatial consistency. Publicly available aerial imagery, including views from Google Earth, was used qualitatively to guide the placement and appearance of sidewalks, traffic infrastructure, and surrounding ground surfaces, ensuring consistency with the real-world corridor. Figure~\ref{fig_visual_comparison} shows representative CARLA UE5 views alongside corresponding locations from Google Earth, illustrating how aerial imagery was used for environmental design.

\subsection{Import into VISSIM}
The finalized HD road network generated in RoadRunner was imported into VISSIM for microscopic traffic simulation. The import preserved lane-level road geometry, intersection topology, and connectivity, enabling direct reuse of the corridor layout across simulators. After import, the road network was validated by checking lane counts and widths, link–connector relationships, and continuous vehicle paths across junctions. Turning movements and priority rules were also reviewed to ensure correct maneuver representation without topological conflicts. 

VISSIM 2026 also imports the pedestrian infrastructure, including sidewalks. As sidewalk geometry was generated automatically, small height adjustments were applied to align the non-planar road surface introduced by LiDAR-derived elevation. These adjustments were limited to pedestrian areas and did not affect road geometry and lane configuration. No scaling or geometric transformation was applied during import, ensuring geometric consistency between VISSIM and CARLA and enabling accurate spatial alignment during co-simulation.

\subsection{Traffic Configuration in VISSIM}
Within the proposed co-simulation framework, VISSIM is the authoritative source of traffic behavior and control logic, ensuring that vehicle, pedestrian, and signal dynamics are governed by a single traffic model. This design eliminates ambiguity in traffic state ownership and enables reproducible traffic scenarios across co-simulation runs. Traffic demand was configured at the corridor level, including definition of vehicle routes, flow rates, and vehicle type distributions representative of typical urban traffic conditions. Pedestrians were also configured to interact with vehicular traffic at intersections and crosswalks, allowing multimodal interactions to be coordinated entirely within VISSIM. Intersection control logic, including signal phasing, priority rules, and permitted turning movements, was defined within VISSIM. During runtime, VISSIM updates the states of all traffic participants and control devices at each simulation step. The resulting state information, including vehicle and pedestrian pose, velocity, and traffic signal phase, is transmitted to the co-simulation interface and propagated to CARLA. As a result, CARLA renders and senses behavior fully dictated by VISSIM, without duplicating or approximating traffic logic.

By centralizing traffic configuration and control within VISSIM, the framework supports systematic experimentation with traffic demand, pedestrian behavior, and signal control strategies while maintaining geometric and temporal consistency across simulators. This separation of responsibilities is a key design feature for smart-corridor research.

\section{CARLA–VISSIM Co-Simulation Framework}

\subsection{System Architecture and Deployment}
The CARLA–VISSIM co-simulation framework integrates a high-fidelity 3D driving simulator with a microscopic traffic simulator across heterogeneous operating systems. Because PTV VISSIM is available only for Windows, while CARLA UE5 with GPU acceleration was executed natively on Ubuntu, VISSIM was deployed inside a Windows virtual machine (VM) on the same host. TCP round-trip measurements across the Ubuntu–VM boundary showed an average latency of 0.073 ms, remaining well below the 0.05 s synchronous simulation step. The system follows a distributed server–client architecture in which CARLA UE5 and VISSIM execute on the server, while a separate client machine runs the driving agent and Python control logic through CARLA.

Communication between CARLA and VISSIM uses a lightweight TCP client–server architecture, with CARLA operating as the TCP server and VISSIM connecting as the TCP client from the Windows VM. This design choice was motivated by the need for a reliable, ordered, and operating-system-agnostic communication channel across the VM boundary. Messages, including actor spawning, destruction, and state updates, are exchanged across the VM boundary.

Both simulators initialize independently and run their own simulation loops. During startup, CARLA configures the world, loads actor blueprints, enables synchronous mode, and discovers traffic lights for synchronization with VISSIM signal controllers. In parallel, VISSIM loads the network, initializes vehicles and pedestrians, and prepares its signal controllers. Once the TCP connection is established, the co-simulation enters a lockstep execution phase with a shared fixed time step. At each step, each publishes state updates for the actors it controls and receives updates for externally controlled actors. Actor creation and removal are communicated to maintain a consistent shared world. This design ensures CARLA- and VISSIM-controlled agents perceive and respond to each other, avoiding ghosting and enabling coherent interaction.

The proposed architecture enables tightly synchronized, bidirectional co-simulation across heterogeneous platforms in a distributed server–client setup. By colocating CARLA UE5 and PTV VISSIM on a shared server and connecting external agents via a networked CARLA client, the framework supports interactive control with consistent traffic interaction. The lightweight TCP protocol avoids extra middleware or platform-specific shared-memory requirements, simplifying deployment across Linux and Windows while preserving determinism.

\subsection{Synchronous Co-Simulation and Dual Actor Ownership}
The co-simulation runs in synchronous lockstep, with both using independent clocks but sharing the same fixed time step. Each executes its own simulation loop and exchanges state once per step, ensuring neither advances beyond the other. This avoids asynchronous drift while preserving the native execution models of both simulators.

Actor ownership follows a hybrid model. Vehicles use bidirectional ownership: each simulator controls the vehicles it spawns while creating mirrored counterparts in the other. VISSIM-controlled vehicles are recreated in CARLA with motion dictated by VISSIM, while CARLA-controlled vehicles (e.g., driven by an external agent) are mirrored in VISSIM while remaining under CARLA’s control. Pedestrians and traffic signals use unidirectional ownership, controlled solely by VISSIM and mirrored only in CARLA. This approach enables both simulators to contribute actors while maintaining consistent interaction across environments.

Actor lifecycle events are synchronized explicitly. Each simulator detects newly spawned and destroyed actors by comparing current and previous actor sets. Spawn and destroy events are then sent to the peer, which creates or removes the corresponding mirrored actors. This explicit synchronization ensures that actor existence remains consistent across both simulators throughout execution.

At each simulation step, both simulators publish state updates for the actors they control, including pose and motion information. These updates are applied to mirrored actors in the peer simulator during the same step. Importantly, incoming updates do not accumulate in queues or replay historical states. Instead, the system adopts a latest-state-per-actor policy, where only the most recent update for each actor is retained and applied. This prevents backlog accumulation, reduces sensitivity to transient communication delays, and ensures that mirrored actors always reflect the most current available state.

Communication between simulators uses a lightweight TCP protocol with explicit message types for spawning, updating, and destroying actors. Messages are processed asynchronously and stored in thread-safe data structures, while state updates are applied synchronously within each simulator’s main loop. This separation keeps network communication non-blocking without compromising deterministic stepping.

\begin{figure}[!t]
\centering
\includegraphics[width=3.6in]{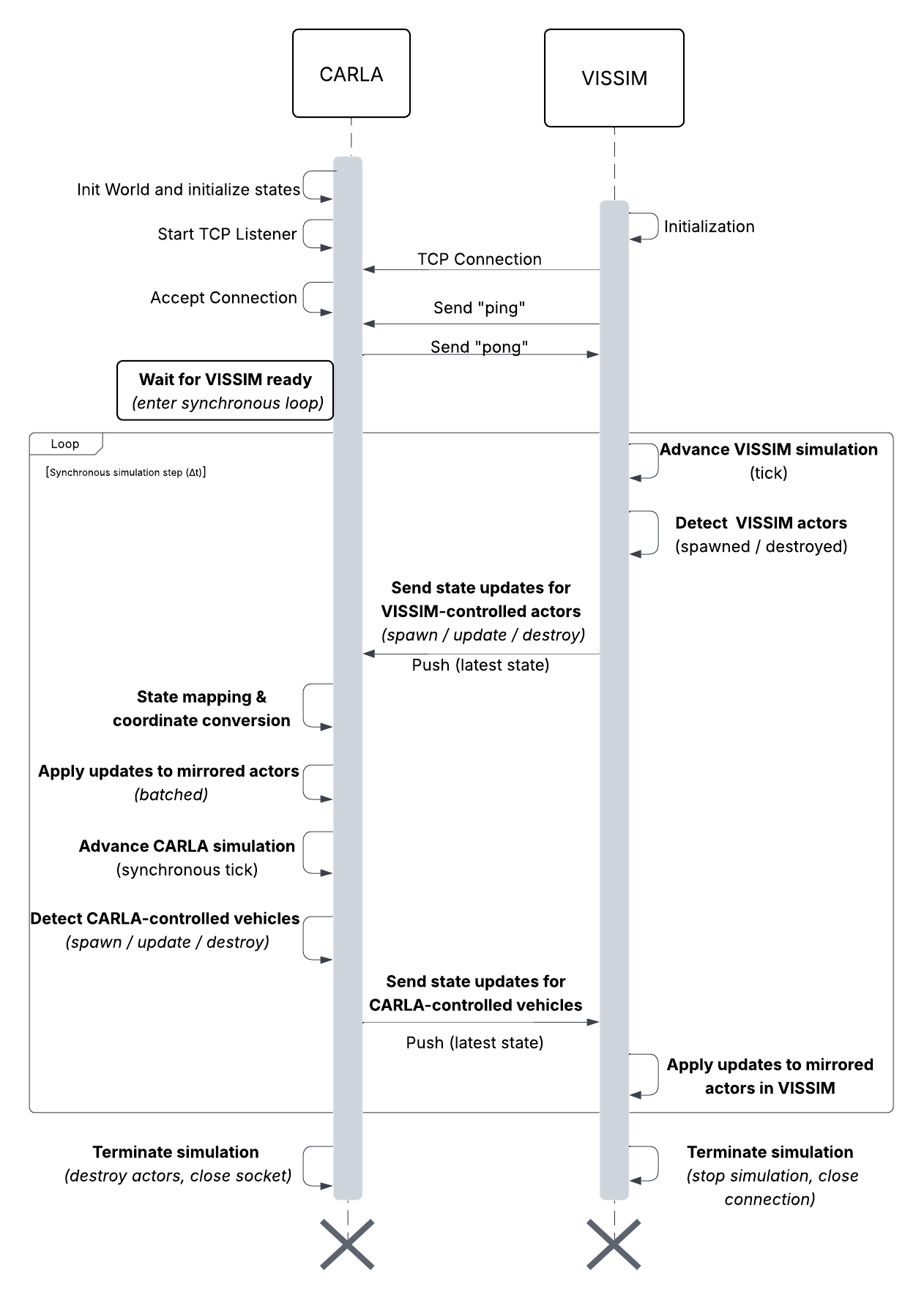}
\caption{\textbf{Timing diagram of the synchronous CARLA–VISSIM co-simulation.} After the initial TCP handshake, CARLA and VISSIM advance with independent clocks but a shared fixed step size ($\Delta t$). At each step, each publishes the latest state of the actors it controls and applies incoming updates to mirrored actors. Actor creation and removal are synchronized explicitly, enabling mixed-authority operation while maintaining consistent state across simulators.}
\label{fig_cosimulation_diagram}
\end{figure}

Overall, the synchronization strategy enables bidirectional, collision-aware co-simulation in which both CARLA and VISSIM contribute controlled actors to a shared traffic environment. Through synchronized time stepping, explicit actor ownership, and a latest-state update policy, the framework achieves stable and consistent co-simulation without a centralized time coordinator or extra synchronization middleware. Figure~\ref{fig_cosimulation_diagram} shows the per-step execution order and bidirectional state exchange within the synchronous co-simulation loop.

\subsection{Cross-Simulator State Mapping and Application}

To maintain a consistent shared traffic environment, the framework maps and applies actor states exchanged between CARLA and VISSIM at every synchronized step. This process is asymmetric in implementation but symmetric in intent: each simulator applies incoming updates only to mirrored actors while preserving full control over its own. 

\subsubsection{Actor state representation and coordinate reconciliation}

Actor state updates include position, orientation, and motion quantities. Because CARLA and VISSIM use different world coordinate conventions, incoming transforms are explicitly converted before application. Mapping from VISSIM to CARLA requires flipping of the lateral axis and adjustment of orientation. These conversions are applied consistently for vehicles and pedestrians before spawning or updating mirrored actors, ensuring spatial alignment across simulators.

\subsubsection{Vehicle state application in CARLA}

For VISSIM-controlled vehicles mirrored in CARLA, state updates are applied using batched CARLA commands rather than direct controls. At each step, CARLA receives updated transforms and velocities, builds a command list of target velocity and transform updates, and executes the batch atomically within the synchronous tick. This approach reflects VISSIM-controlled motion accurately while avoiding per-message blocking.

\subsubsection{Pedestrian state application in CARLA}

Pedestrian synchronization follows a similar pattern but accounts for CARLA’s pedestrian motion model. For VISSIM-controlled pedestrians, transform, velocity, and walker control updates are applied per step, with movement direction derived from the incoming orientation. These updates are issued in the same synchronous batch as vehicles to ensure temporal consistency.

\subsubsection{Actor state application in VISSIM}

For CARLA-controlled actors, CARLA publishes state updates to the VISSIM synchronization loop. These vehicles are spawned in VISSIM when first detected and updated each step using their latest state. VISSIM applies the updates directly to its internal actors, allowing its traffic to perceive and react to these actors.

\subsubsection{Traffic signal mapping and synchronization}

Traffic signal synchronization is handled on the CARLA side through a dedicated mapping layer. During initialization, CARLA discovers traffic light actors, groups them into intersections, and classifies them by direction and role (e.g., main movement vs. protected left). Each light is linked to the corresponding VISSIM controller ID and signal group. During co-simulation, VISSIM signal updates are mapped to the associated CARLA traffic lights and applied synchronously to maintain alignment with VISSIM signal logic and intersection geometry.

\subsubsection{Synchronous application and consistency guarantees}

All actor and signal updates are applied within each simulator’s synchronous step, while network communication remains decoupled from state application. Only the most recent state for each actor is retained, preventing stale updates and maintaining spatial, temporal, and behavioral consistency between CARLA and VISSIM.

\section{Case Study: Urban Corridor Co-Simulation}

\subsection{Corridor Description and Scope}
The case study focuses on a multi-intersection urban corridor along Martin Luther King Jr. Boulevard in Chattanooga, Tennessee. The modeled area includes fifteen intersections, centered on five primary signalized intersections along MLK—Broad Street, Market Street, Georgia Street, Lindsay Street, and Houston Street—plus their directly connected upstream and downstream intersections. All MLK signalized intersections support both vehicle and pedestrian movements, while three westernmost intersections include protected left-turn phases. Pedestrian crossing areas follow real-world markings, including standard crosswalks and intersection-wide pedestrian zones. The remaining corridor-connected intersections are stop-controlled feeder streets that add traffic interaction without centralized signal coordination.

The corridor features closely spaced signalized intersections with concurrent vehicle and pedestrian activity, creating frequent interaction points and overlapping control actions. Each intersection operates with its own signal controller, and all vehicle and pedestrian phases are managed by VISSIM, providing a realistic stress test for cross-simulator synchronization and actor ownership. This corridor-scale testbed enables evaluation of the co-simulation framework under realistic yet controlled urban conditions. The combination of multiple signalized intersections, pedestrian movements, independent signal control, and shared actor authority provides a representative smart-corridor setting for demonstrating the capabilities and robustness of the co-simulation framework.

\subsection{Co-Simulation Scenario Setup}

The co-simulation scenario evaluates the framework under realistic multi-intersection urban traffic conditions along the MLK corridor. CARLA UE5 and VISSIM both operated with a fixed 0.05~s synchronous step, enabling lockstep co-simulation throughout execution. Across repeated runs under peak corridor load (approximately 100 vehicles and 100 pedestrians), the average synchronization processing overhead remained under 25~ms per step. Despite transient variations during actor spawning and synchronization updates, the framework maintained stable lockstep co-simulation under the fixed 0.05~s synchronization step.

Traffic participants included VISSIM-controlled vehicles, pedestrians, and traffic signals, together with a CARLA-controlled ego vehicle operated through a CARLA client. At peak demand, about 100 vehicles and 100 pedestrians were active within the corridor. Pedestrian actors traversed sidewalks between intersections and performed signalized crossings at MLK intersections, with all pedestrian behavior governed exclusively by VISSIM’s microscopic logic.

Traffic signal operation was defined entirely in VISSIM. Each signalized intersection along the MLK corridor was controlled by an independent signal controller managing both vehicle and pedestrian phases. Signal states generated by VISSIM were mirrored in CARLA at runtime, allowing CARLA-rendered traffic lights to reflect the same phase transitions experienced by VISSIM-controlled traffic participants.

The CARLA environment visualized the corridor-scale scene and enabled interactive driving of the ego vehicle, while background traffic, pedestrian movement, and signal timing were governed by VISSIM. The CARLA-controlled ego vehicle interacted with VISSIM-controlled traffic through the bidirectional state exchange described earlier, allowing evaluation of vehicle–pedestrian interaction, traffic signal synchronization, and cross-simulator consistency under realistic conditions. Each simulation run covered five minutes, providing multiple signal cycles and pedestrian crossings across the corridor and offering sufficient scope to assess synchronization stability, interaction fidelity, and consistent behavior across all intersections without unnecessary complexity.

\subsection{Demonstrated Capabilities and Qualitative Results}

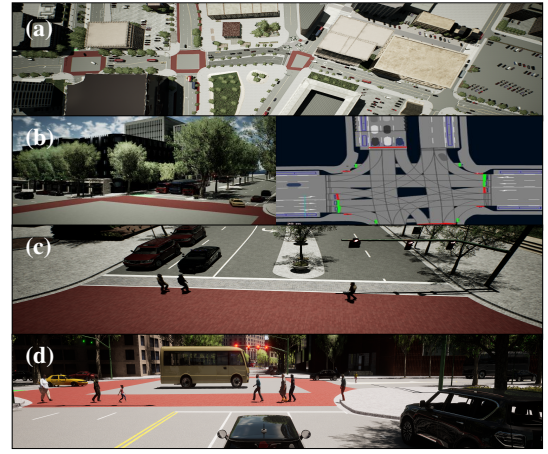
\begin{figure}[t]
\centering
\setlength{\fboxsep}{0pt}
\setlength{\fboxrule}{0pt}

\newcommand{\FixedPic}[3]{%
        \begin{tikzpicture}
            \node[anchor=south west, inner sep=0pt] (img) at (0,0)
                {#1};

            \draw[black, line width=0.3pt]
                (img.south west) rectangle (img.north east);

            \node[anchor=north west,font=\bfseries\small,text=white]
                at (0.05,1.47) {#2};
        \end{tikzpicture}
}

\begin{subfigure}[t]{0.38\textwidth}
    \centering
    \FixedPic{\includegraphics[width=7cm,height=1.55cm]{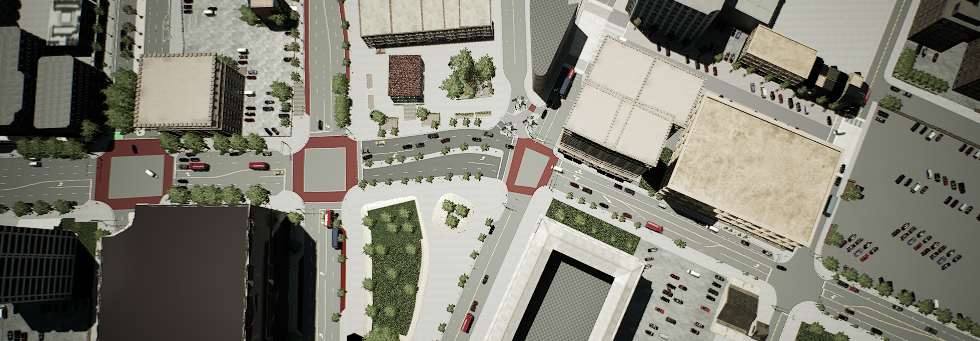}}{(a)}{}
    \refstepcounter{subfigure}
    \label{fig_corridor_overview}
\end{subfigure}\\[-5mm]
\begin{subfigure}[t]{0.38\textwidth}
    \centering
    \FixedPic{%
        \includegraphics[width=3.49cm,height=1.5cm]{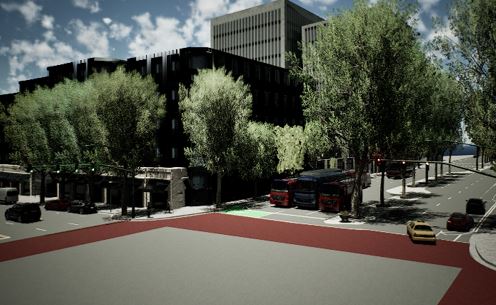}%
        \includegraphics[width=3.51cm,height=1.5cm]{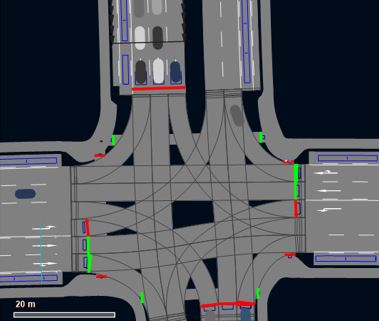}%
    }{(b)}{}
    \refstepcounter{subfigure}
    \label{fig_signal_mirroring}
\end{subfigure}\\[-5mm]
\begin{subfigure}[t]{0.38\textwidth}
    \centering
    \FixedPic{\includegraphics[width=7cm,height=1.5cm]{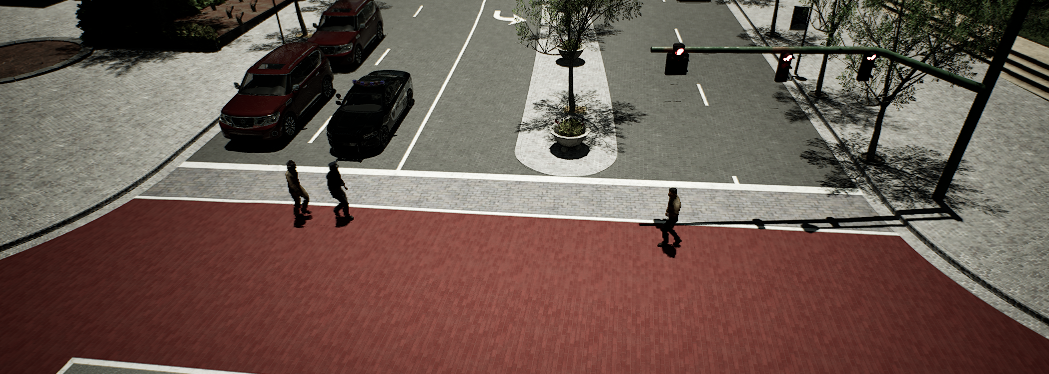}}{(c)}{}
    \refstepcounter{subfigure}
    \label{fig_vehicle_pedestrian_interaction}
\end{subfigure}\\[-5mm]
\begin{subfigure}[t]{0.38\textwidth}
    \centering
    \FixedPic{\includegraphics[width=7cm,height=1.5cm]{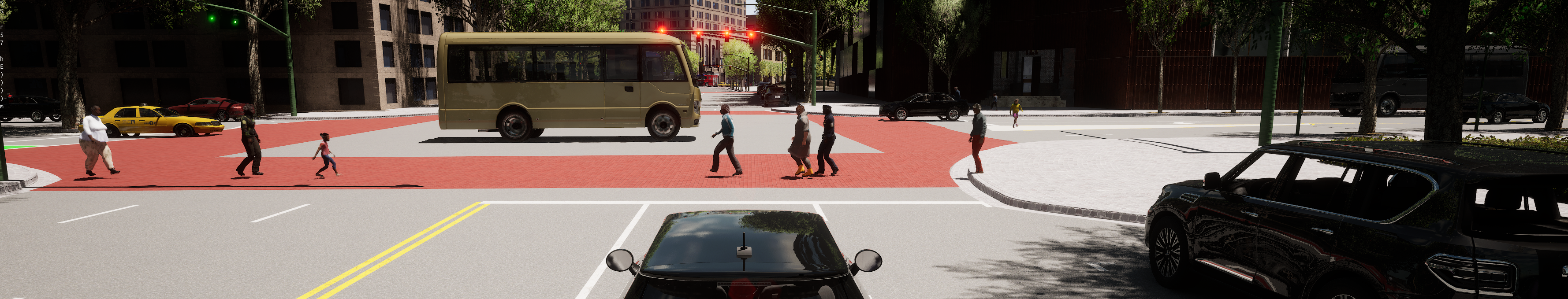}}{(d)}{}
    \refstepcounter{subfigure}
    \label{fig_mixed_authority_ego_vehicle}
\end{subfigure}

\caption{\textbf{Qualitative results from the CARLA–VISSIM co-simulation.}
(a) Corridor overview, 
(b) signal mirroring, 
(c) vehicle–pedestrian interaction, and
(d) mixed-authority operation with a human-driven ego vehicle.}
\label{fig_corridor_overview_total}
\end{figure}

Figure~\ref{fig_corridor_overview} presents the simulated corridor in CARLA UE5, illustrating a multi-intersection environment used for qualitative evaluation. The visualization shows multiple closely spaced signalized intersections and adjacent connecting streets, featuring active vehicles and pedestrians that produce overlapping interactions and frequent signal-controlled decisions. This setup provides a visually and operationally representative urban environment in which cross-simulator synchronization, actor ownership, and traffic control consistency can be qualitatively assessed under realistic multi-intersection conditions.

Figure~\ref{fig_signal_mirroring} shows traffic signal synchronization at the Broad Street intersection, where vehicle and pedestrian phases alternate to ensure exclusive right-of-way. Vehicle red phases align with the pedestrian greens, and vice versa. VISSIM generates signal states and transmits them to CARLA each step, where they are applied to the corresponding traffic lights. The resulting CARLA states match the VISSIM phases, confirming that vehicle stops, pedestrian crossing permissions, and subsequent changes are mirrored consistently. With VISSIM as the authoritative controller while rendering signal states in CARLA, the framework supports perception-oriented evaluation of signalized intersections under traffic control logic.

Figure~\ref{fig_vehicle_pedestrian_interaction} illustrates vehicle–pedestrian interaction at the Broad Street intersection under signalized control. During pedestrian green phases, pedestrians cross using designated pedestrian areas while all approaching vehicles remain stopped at the stop lines. This behavior is consistently observed across simulation runs and reflects the complementary vehicle–pedestrian signal logic defined in VISSIM. Pedestrians move only during authorized crossing phases, and vehicles resume movement once the pedestrian phase ends. This interaction demonstrates that pedestrian right-of-way and vehicle compliance are preserved across simulators, with VISSIM governing the logic and CARLA accurately rendering the behavior. These observations confirm that the co-simulation framework supports realistic and synchronized vehicle–pedestrian interaction at signalized urban intersections.

Figure~\ref{fig_mixed_authority_ego_vehicle} illustrates mixed-authority interaction at a signalized intersection, where a human-driven CARLA ego vehicle operates alongside VISSIM-controlled traffic. The ego vehicle is driven through a CARLA client interface with a steering wheel and pedals, while surrounding vehicles, pedestrians, and signals follow VISSIM's logic. This setup enables direct interaction between human-driven behavior and VISSIM-governed traffic within a shared environment. The ego vehicle follows traffic signals and pedestrian right-of-way while interacting naturally with VISSIM-controlled vehicles at the intersection. The ego vehicle stops for red lights and pedestrian phases, and VISSIM-controlled traffic simultaneously perceives and reacts to the ego vehicle as part of the shared traffic stream. These observations show that the co-simulation framework supports seamless integration of human-driven agents with simulator-controlled traffic, enabling realistic mixed-authority interaction at corridor-scale signalized intersections.

\section{Corridor-Scale Deployment Insights}

Deploying the CARLA–VISSIM co-simulation framework on a multi-intersection segment of MLK Street revealed several practical insights not evident from simulator integration alone. Although the modeled environment does not cover an entire city, the corridor-scale deployment was sufficient to expose synchronization challenges, interaction complexity, and design trade-offs relevant to smart-corridor applications.

\noindent\textbf{Corridor Scale as an Effective Stress Test.}
 A corridor with roughly fifteen intersections, including five closely spaced signalized intersections with pedestrian phases, was sufficient to stress the co-simulation framework in meaningful ways. Concurrent signal controllers, continuous vehicle flow from stop-controlled side streets, and frequent pedestrian crossings created overlapping control events along the corridor. This setup exercised synchronization, actor ownership, and signal-state consistency more effectively than isolated intersection tests, while avoiding the unnecessary complexity of full city-scale modeling. The scale also aligns with how real-world smart-corridor pilots are typically deployed and evaluated.

\noindent\textbf{Importance of Explicit Actor Ownership.}
One key design decision validated during deployment was assigning behavioral authority to the simulator that spawns each actor. This prevents ambiguous control when human-driven and simulator-controlled agents interact at signalized intersections. Mirroring only state information—rather than shared or blended control—was essential for maintaining consistent behavior across simulators. This was particularly important for pedestrian agents, whose motion and right-of-way were governed entirely by VISSIM while being rendered in CARLA.

\noindent\textbf{Stability Through Latest-State Synchronization.}
The latest-state-per-actor update policy was key to maintaining stable operation under corridor-scale traffic density. With 200 active vehicles and pedestrians during peak periods, state updates occurred across multiple intersections. Keeping only the most recent state for each actor avoided backlog accumulation and reduced sensitivity to communication delays. During steady-state operation, no ghosting, duplicates, or noticeable desynchronization appeared, even as multiple signal phases and pedestrian crossings occurred along the corridor.

\noindent\textbf{Initialization and Operational Observations.}
While steady-state operation was stable, transient artifacts occasionally appeared during initialization or after long execution periods. These artifacts were resolved by restarting the session, underscoring the need to separate initialization from runtime synchronization in practical deployments. Once the system entered steady-state lockstep execution, behavior across CARLA and VISSIM remained consistent throughout the runs.

\noindent\textbf{Implications for Smart-Corridor Applications.}
This deployment shows that the MLK urban corridor provides an effective scale for evaluating synchronization, actor authority, and signal-state consistency in our CARLA–VISSIM integration. Integrating human-driven vehicles, microscopic traffic control, and high-fidelity visualization in a single synchronized environment enables experiments that would be difficult to achieve in either simulator alone. These insights indicate that full city-scale modeling is not a prerequisite for meaningful smart-corridor evaluation and that targeted corridor deployments can provide value for both research and planning studies.

\section{Conclusion}

This paper presented a corridor-scale co-simulation framework integrating CARLA UE5 and PTV VISSIM to support realistic urban traffic experimentation with vehicles, pedestrians, and signalized intersections. By decoupling microscopic control logic from high-fidelity visualization and interaction, the framework allows each simulator to operate within its strength while maintaining a synchronized traffic environment.

The MLK Street case study showed that corridor-scale deployments are sufficient to stress synchronization mechanisms, actor ownership policies, and signal-state consistency, providing practical insights for smart-corridor applications. Qualitative results showed consistent signal mirroring, realistic vehicle–pedestrian interaction, and stable mixed-authority operation involving a human-driven ego vehicle, with no ghosting or desynchronization during steady-state execution.

Overall, these findings show that tightly synchronized co-simulation across the fifteen-intersection MLK corridor provides a practical foundation for studying mixed vehicle–pedestrian interaction, signal consistency, and human-in-the-loop driving using CARLA and VISSIM.

\section{Acknowledgment}
The authors thank Firas Elhag for his valuable assistance with the UE5 traffic light setup and related logic within the co-simulation environment.

\bibliographystyle{IEEEtran}
\bibliography{ref}

\end{document}